%% file: a_arxiv_version.tex
\documentclass[10pt,twocolumn,letterpaper]{article}

\usepackage[pagenumbers]{iccv} 

\usepackage{multirow}
\usepackage{multicol}
\usepackage{graphicx}
\usepackage{subfloat}
\usepackage{amsmath}
\usepackage{amssymb}
\usepackage{booktabs}


\definecolor{iccvblue}{rgb}{0.21,0.49,0.74}
\usepackage[pagebackref,breaklinks,colorlinks,allcolors=iccvblue]{hyperref}

\def\paperID{13444} 
\def\confName{ICCV}
\def\confYear{2025}

\title{DiGA3D: Coarse-to-Fine Diffusional Propagation of Geometry and Appearance for Versatile 3D Inpainting}

\author{Jingyi Pan$^{1}$ \quad Dan Xu$^{2}\footnotemark[1]~$ \quad Qiong Luo$^{1,2}\footnotemark[1]~$ 
\\
$^{1}$The Hong Kong University of Science and Technology (Guangzhou)
\\
$^{2}$The Hong Kong University of Science and Technology\\ 
{\tt\small jpan305@connect.hkust-gz.edu.cn, \{danxu, luo\}@cse.ust.hk}\\
\vspace{-30pt}}

\begin{document}
\maketitle
\footnotetext[1]{Corresponding authors.} 
\input{sec/0_abstract}

\input{sec/1_intro}
\input{sec/2_related_work}

\input{sec/3_methods}

\input{sec/4_experiments}
\input{sec/5_conclusion}
\input{sec/6_suppl}

\input{sec/7_ack}
\clearpage
{
    \small
    \bibliographystyle{ieeenat_fullname}
    \bibliography{main}
}

\end{document}

%% file: sec/0_abstract.tex
\begin{abstract}
Developing a unified pipeline that enables users to remove, re-texture, or replace objects in a versatile manner is crucial for text-guided 3D inpainting. However, there are still challenges in performing multiple 3D inpainting tasks within a unified framework: 1) Single reference inpainting methods lack robustness when dealing with views that are far from the reference view; 2) Appearance inconsistency arises when independently inpainting multi-view images with 2D diffusion priors; 3) Geometry inconsistency limits performance when there are significant geometric changes in the inpainting regions. To tackle these challenges, we introduce \textbf{DiGA3D}, a novel and versatile 3D inpainting pipeline that leverages diffusion models to propagate consistent appearance and geometry in a coarse-to-fine manner. First, DiGA3D develops a robust strategy for selecting multiple reference views to reduce errors during propagation. Next, DiGA3D designs an Attention Feature Propagation (AFP) mechanism that propagates attention features from the selected reference views to other views via diffusion models to maintain appearance consistency. Furthermore, DiGA3D introduces a Texture-Geometry Score Distillation Sampling (TG-SDS) loss to further improve the geometric consistency of inpainted 3D scenes.
Extensive experiments on multiple 3D inpainting tasks demonstrate the effectiveness of our method. The project page is available at \href{https://rorisis.github.io/DiGA3D/}{HERE}.


\end{abstract}

%% file: sec/1_intro.tex
\section{Introduction}
\label{sec:intro}

Recent advances in 3D representations \cite{mildenhall2021nerf, wang2021neus, kerbl20233d} and text-to-image (T2I) diffusion models have led to significant progress in novel view synthesis (NVS) and 3D generation, demonstrating substantial potential for applications in areas such as VR/AR and the Metaverse.
Despite these advances, 3D inpainting, particularly the development of unified pipelines for various 3D inpainting tasks, remains a relatively less-studied area. 

\begin{figure}[t]
    \centering
    \includegraphics[width=\linewidth]{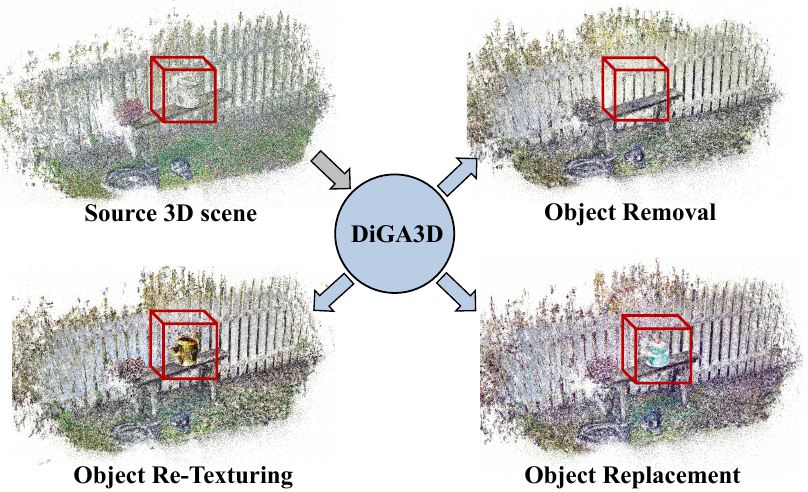}
    \vspace{-15pt}
    \caption{\small DiGA3D is a versatile 3D inpainting framework guided by text prompts, supporting multiple inpainting tasks including object replacement, removal, and re-texturing, \etc.}
    \label{fig:fig1}
    \vspace{-15pt}
\end{figure}

Although several methods~\cite{liu2022nerf, mirzaei2023reference, wang2024innerf360, chen2024gaussianeditor, bartrum2024replaceanything3d} have explored unified pipelines for versatile 3D inpainting, they still face some challenges: First, some methods~\cite{liu2022nerf, mirzaei2023reference} rely on a single reference image to guide the inpainting process, which heavily depends on the quality of the single reference image and often leads to texture degradation when views are far from the reference view; Second, some methods~\cite{mirzaei2023spin, wang2024innerf360} struggle to maintain multi-view appearance consistency as they independently inpaint the constituent images using 2D inpainters. Although they utilize perceptual loss~\cite{zhang2018unreasonable} to optimize the views and address these inconsistencies subsequently, they are inadequate when the appearance of the inpainted views differs perceptually; Third, existing methods frequently suffer from inconsistent geometry, leading to issues such as multi-facet artifacts. Although some approaches~\cite{mirzaei2023reference, wang2024gscream, bartrum2024replaceanything3d} attempt to address geometric inconsistencies by incorporating depth maps generated by monocular depth estimators, they often rely on depth maps that are inconsistent across multiple views. This limitation becomes particularly evident when inpainting regions require significant geometric changes.

To address these challenges, we introduce DiGA3D, a novel and versatile 3D inpainting pipeline with a coarse-to-fine manner that utilizes 3D Gaussian Splatting (3DGS) to leverage diffusion priors for propagating appearance and geometry across multiple views. To mitigate the multi-view bias when guided by a single reference image, we develop a robust strategy for selecting multiple reference views to reduce the propagation errors caused by these reference views. In the coarse stage, we propose a multi-view inpainting scheme by propagating attention features from reference views to other views through the latent space within diffusion models, thereby implicitly ensuring the appearance consistency of multi-view images. In the fine stage, we design a Texture-Geometry-guided Score Distillation Sampling (TG-SDS) loss as a geometric regularization. This involves using warped texture images and depth maps from reference views as conditional inputs for multi-control diffusion models~\cite{zhang2023adding}. 
Furthermore, this loss explicitly and controllably propagates textural and geometric information from the selected reference views, further enhancing both the appearance and geometry in the 3D inpainting process. Thus, our method offers a coarse-to-fine pipeline that can effectively bridge consistent 2D appearance and 3D geometry, enabling versatile 3D inpainting.

Extensive experiments across various 3D inpainting tasks, such as object removal, object re-texturing, and object replacement in diverse scenes, demonstrate the effectiveness of our method, as depicted in Fig.~\ref{fig:fig1}. In summary, our key contributions can be outlined as follows:
\begin{itemize} 
\item We introduce DiGA3D, a versatile 3D inpainting pipeline that leverages diffusion models to consistently propagate appearance and geometry in a coarse-to-fine manner.
\item We develop an Attention Feature Propagation (AFP) mechanism within the 2D inpainter to achieve coarsely consistent inpainting results. 
\item We propose a Texture-Geometry-guided Score Distillation Sampling (TG-SDS) optimization loss to enhance the geometric and appearance consistency across all views. 
\item Extensive experiments on several 3D inpainting tasks demonstrate the effectiveness of our method. 
\end{itemize}

%% file: sec/2_related_work.tex
\section{Related Work}
\label{sec:formatting}

\begin{figure*}[t]
    \centering
    \includegraphics[width=\linewidth]{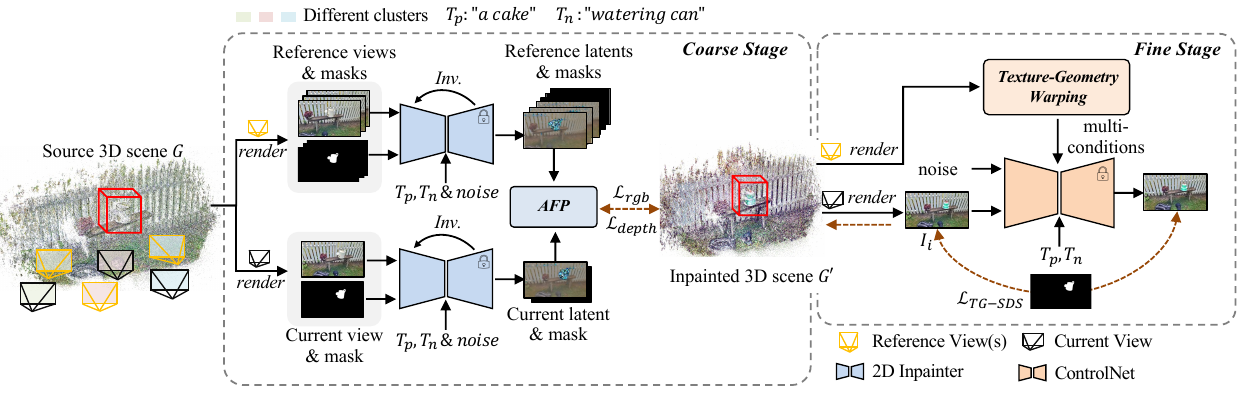}
    \vspace{-23pt}
    \caption{\small \textbf{Our proposed framework.} Before performing 3D inpainting, we first calculate the camera pose using COLMAP~\cite{schonberger2016structure} and extract masks from mask prompts $T_m$. We then apply k-means clustering to group the views based on their camera centers and select the views closest to the cluster centers as the reference views. In the coarse stage, we employ DDIM Inversion~\cite{song2021denoising} to generate deterministic latents, which are then used to produce coarsely consistent inpainting results with a 2D inpainter equipped with the AFP module. In the fine stage, we utilize ControlNet~\cite{zhang2023adding}, leveraging texture and depth images as conditions, to further refine the inpainted 3D scene by TG-SDS loss. In this scene, we designate $T_p$ as ``a cake" and $T_n$ as ``watering can" to replace the watering can with a cake.}
    \vspace{-8pt}
    \label{fig:framework}
\end{figure*}

\noindent\textbf{2D Inpainting.}
Image inpainting aims to restore missing regions in masked regions while preserving rich textures and structural integrity. Early classic methods primarily involved copying textures from known areas into unknown ones~\cite{efros1999texture}. In recent years, learning-based approaches have made significant advancements in this field. For instance, LaMa~\cite{suvorov2022resolution} demonstrates a strong ability to fill large missing areas using fast Fourier convolutions. Additionally, developments in diffusion models~\cite{rombach2022high} have resulted in remarkable improvements, with models like SD-inpainter~\cite{rombach2022high} producing diverse inpainting results for masked regions. However, many of these approaches necessitate fine-tuning for specific downstream tasks. 
Furthermore, the scope of image inpainting has been extended to video inpainting. Some methods~\cite{zhou2021transfill, cao2024leftrefill} utilize one or more reference inpainting images to propagate content throughout the entire video. Other approaches~\cite{gao2020flow, li2022towards, zhou2023propainter} leverage optical flow as a prior to capture motion, ensuring temporal consistency in the inpainting process. 
Unlike traditional image or video inpainting methods~\cite{rombach2022high}, which rely on text prompts to describe inpainting regions, expanding to 3D inpainting presents challenges for complex or 360-degree scenes, where backgrounds are hard to summarize with a single description. Fortunately, PowerPaint~\cite{zhuang2023task} offers a unified framework that manages multiple inpainting tasks using task-specific prompts, enabling versatile 3D inpainting within a unified pipeline.

\noindent\textbf{3D Inpainting in NeRF and 3DGS.}
With the rapid advancement of neural scene representations~\cite{mildenhall2021nerf, wang2021neus, kerbl20233d}, there is an increasing demand for 3D inpainting~\cite{mirzaei2023spin, chen2024mvip, weber2024nerfiller, liu2024infusion, cao2024mvinpainter}. The objective of 3D inpainting is to fill in missing regions within a 3D scene, such as removing objects and generating realistic textures and geometries to complete the affected areas.
These methods can be broadly categorized into those that utilize diffusion models and those that do not. Some approaches~\cite{yang2021self, qin2024langsplat, zhou2024feature} leverage CLIP~\cite{radford2021learning} or DINO features~\cite{caron2021emerging} to capture 3D semantics, enabling targeted inpainting of specific regions based on the characteristics of the 3D representations. In contrast, diffusion-guided methods often rely on a reference image to propagate texture and geometry from that reference view across all views~\cite{liu2022nerf, mirzaei2023reference, wang2024gscream, lu2024view}. Other approaches~\cite{liu2024infusion, cao2024mvinpainter} enhance the consistency and plausibility of inpainting results by fine-tuning diffusion models using depth or optical flow priors. 
Unlike these diffusion-guided approaches, we utilize training-free diffusion models for various 3D inpainting tasks. Our method employs attention feature propagation within the diffusion models and explicitly incorporates texture and geometry information as conditions to further ensure consistency in both appearance and geometry.

%% file: sec/3_methods.tex
\section{Method}

\subsection{Preliminary}
\textbf{3D Gaussian Splatting.}
Gaussian Splatting \citep{kerbl20233d} is a point-based 3D representation method. 
Each \emph{Gaussian ellipse} is defined by a color $c$ represented with spherical harmonics coefficients, an opacity $o$, a position center $\mu$, and a \emph{covariance matrix} $\Sigma$. The Gaussian ellipse is calculated as $G(x) = e^{-\frac{1}{2}x^{T}\Sigma^{-1}x}$, where $x$ is the displacement from the center $\mu$. The covariance matrix $\Sigma$ can be decomposed into a \emph{rotation matrix} $R$ and a \emph{scaling matrix} $S$ for differentiable optimization: $\Sigma = RSS^{T}R^{T}$. During the rendering process, 3D Gaussians are projected onto 2D planes using a \emph{splatting} operation \citep{zwicker2001ewa}, which positions the Gaussians using a new covariance matrix $\Sigma^{'}$ in camera coordinates, defined as $\Sigma^{'} = JW\Sigma W^{T}J^{T}$. Here, $J$ is the Jacobian of the affine approximation of the projective transformation, and $W$ is the given viewing transformation matrix. The rendering results $C$ at a pixel is achieved by approximating the projection of a 3D Gaussian along the depth dimension onto the pixel:
{\setlength\abovedisplayskip{2pt}
\setlength\belowdisplayskip{2pt}
\begin{equation}
    C = \sum_{i\in N}c_i\alpha_i\prod_{j=1}^{i-1}(1-\alpha_j),
\end{equation}}
where $N$ is the set of ordered points that project onto the pixel, ensuring coherent rendering of overlapping Gaussians.

\noindent\textbf{Score Distillation Sampling.}
Text-to-3D has seen significant advancements by optimizing a 3D representation using a 2D pre-trained image diffusion prior $\epsilon_{\phi}$, based on Score Distillation Sampling (SDS)~\cite{poole2023dreamfusion}. The diffusion model $\phi$ is pre-trained to predict sampled noise $\epsilon_{\phi}(x_t; t, y)$ that adds noise to the image $x$ at timestep $t$, conditioned on the text embeddings $y$. By rendering a random view through a differentiable renderer $g(\cdot)$, SDS updates the parameter $\theta$ by randomly selecting timesteps $t \sim \mathcal{U}(t_{\tt{min}}, t_{\tt{max}})$ and forwarding $x = g(\theta)$ with noise $\epsilon \sim \mathcal{N}(0, I)$ to compute the gradient as follows:
\begin{align}\label{eq:sds}
    \nabla_\theta \mathcal{L}_{\tt{SDS}}(\theta) = \mathbb{E}_{t,\epsilon}\left[ w(t) \big(\epsilon_\phi(x_t; y,t) - \epsilon\big)\frac{\partial x}{\partial \theta} \right].
\end{align}

\begin{figure}
    \centering    \includegraphics[width=\linewidth]{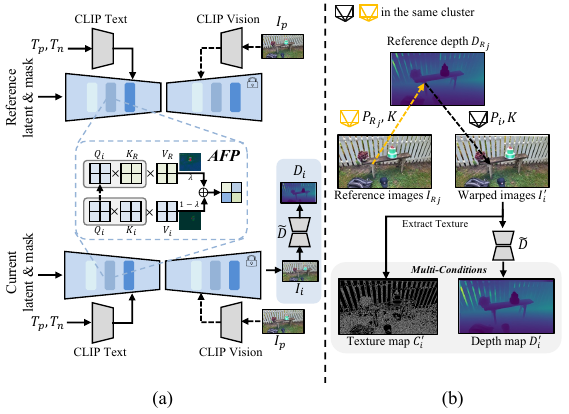}
    \vspace{-19pt}
    \caption{\small (a) The illustration of the proposed Attention Feature Propagation (AFP). The outputs of AFP are the inpainted image $I_i$ and the depth map $D_i$ estimated by the monocular depth estimator~\cite{ranftl2021vision} $\tilde{D}$. (b) The workflow of our designed texture-geometry warping module. The outputs of texture-geometry warping are the texture map $C'_i$ and the depth map $D'_i$.}
    \vspace{-15pt}
    \label{fig:method}
\end{figure} 

\subsection{Problem formulation and overview}
We define the problem of versatile 3D inpainting using 3DGS as follows: Given a pretrained 3D Gaussians $G$, a positive prompt $T_p$, a negative prompt $T_n$ describing the inpainting target, and a mask prompt $T_m$ to guide the Language-based Segment Anything model (Lang SAM)~\citep{kirillov2023segment} in selecting specific inpainting regions, our objective is to inpaint the 3D Gaussians based on these text prompts.

As illustrated in Fig.~\ref{fig:framework}, we use a coarse-to-fine strategy for versatile and view-consistent 3D inpainting from multi-view images. Prior to the 3D inpainting process, we initialize the camera poses for the 3D scene and apply K-means clustering~\cite{lloyd1982least} to group the multi-view images based on camera centers derived from COLMAP~\cite{schonberger2016structure}. We then choose the views closest to the cluster centers as reference views.
In the coarse stage, we employ DDIM inversion and the Attention Feature Propagate (AFP) module, allowing attention features to propagate from reference views to other views, thereby optimizing a coarsely view-consistent 3D Gaussians (see Sec.~\ref{sec:consistent}).
In the fine stage, we leverage the TG-SDS loss as geometry regularization to improve both geometry and texture of the inpainted 3D scenes (see Sec.~\ref{sec:sds}). 
The overall loss functions are shown in Sec.~\ref{sec:optimization}.

\begin{figure}
    \centering
    \includegraphics[width=\linewidth]{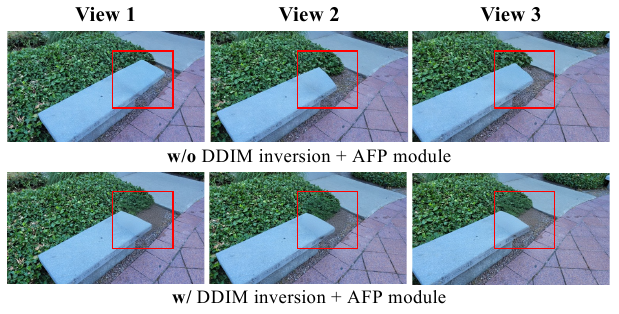}
    \vspace{-15pt}
    \caption{\small Illustration of the multi-view consistent image inpainting with DDIM inversion and the AFP module in Sec.~\ref{sec:consistent}.}
    \vspace{-10pt}
    \label{fig:example}
\end{figure}
\subsection{Multi-view Consistent Image Inpainting} \label{sec:consistent}
Achieving high-quality inpainted 3D scenes is a challenging task because existing 2D inpainters~\cite{rombach2022high, zhang2023adding, zhuang2023task} struggle to produce consistent multi-view inpainting results.
Drawing inspiration from video editing and inpainting methods~\cite{wu2023tune, liu2024video}, we design the Attention Feature Propagation (AFP) strategy. This strategy aims to implicitly propagate attention features from reference views to other views within the latent space of a 2D inpainter. Prior to employing AFP, we introduce a robust strategy for selecting the reference views.

\noindent\textbf{Reference Views Selection.} To ensure that the selected reference views capture the majority of appearance and geometric information across the entire scene, we utilize K-means clustering to group all views based on their camera centers, which are determined through pose estimation using COLMAP~\cite{schonberger2016structure}. As shown in Fig.~\ref{fig:framework}, this process results in $K$ clusters within the scene. We then select the views closest to the cluster centers as reference views. This straightforward yet effective method enables us to choose reference views that can establish relationships with surrounding views and minimize warping errors.

\noindent\textbf{DDIM Inversion.}
In the coarse stage, as depicted in Fig.~\ref{fig:framework}, to facilitate the generation of 3D-consistent coarse appearances, we apply DDIM inversion on rendered images $\hat{I}$ from source 3D Gaussians and masks extracted by Lang SAM~\cite{kirillov2023segment} to derive intermediate deterministic latents $z^t$ from the 2D inpainter.

\noindent\textbf{Attention Features Propagation (AFP).}
After deriving the deterministic latents via DDIM inversions, we leverage these latents to enhance multi-view appearance consistency. To propagate the inpainted appearance from reference views, we first integrate a self-attention mechanism~\citep{wu2023tune} to extract attention features from each view, as shown in Fig.~\ref{fig:method} (a). Subsequently, we employ a cross-attention mechanism to inject reference attention features into the inpainting process of other views. The self-attention mechanism is described as:
{\setlength\abovedisplayskip{2pt}
\setlength\belowdisplayskip{2pt}
\begin{equation}
\begin{split}
    {\rm Attn(Q_i, K_i, V_i)} = {\rm Softmax}(\frac{Q_i K_i}{\sqrt{d}})V_i,
\end{split}
\label{eq:attn_k}
\end{equation}}
where $Q_i$, $K_i$, and $V_i$ represent the Query, Key, and Value features obtained from linear projections of the self-attention mechanism for latents $z^t$ of each view, with $d$ acting as a scaling factor. 

Furthermore, we utilize cross-attention to incorporate the attention features from reference views into the attention features of other views:
{\setlength\abovedisplayskip{2pt}
\setlength\belowdisplayskip{2pt}
\begin{equation}
\begin{split}
    {\rm Attn'_i} &= \lambda_a \cdot \frac{1}{N_k}\sum_{i=0}^{N_k} Attn(Q_i, K_r, V_r) \\
    &+ (1- \lambda_a) \cdot Attn(Q_i, K_i, V_i),
\end{split}
\label{eq:attn}
\end{equation}}
where $\lambda_a \in [0, 1]$, and $N_k$ represents the number of reference views selected from K-means. 
To further assist in improving appearance consistency, we encode the already inpainted image $I_p$ within the multi-view sequence using the CLIP Vision model~\cite{radford2021learning} and integrate the image embeddings into the residual blocks of the U-Net.
Next, we decode inpainted latents to produce coarsely consistent inpainted results for training the 3D Gaussians.
\subsection{Texture-Geometry Guided SDS Loss}\label{sec:sds}

By optimizing 3D Gaussians using these inpainted images, we can generate coarsely inpainted 3D scenes. While we have achieved relatively consistent inpainting results, as shown in Fig.~\ref{fig:example}, these results might lack the essential geometric information necessary for 3D inpainting. 
Furthermore, the AFP module facilitates the propagation of attention features from reference views to other views, aiding in enhancing appearance consistency implicitly. 
However, this approach may not comprehensively address all detail inconsistencies. To further alleviate artifacts in 3D inpainting, we propagate geometry and texture details in an explicit and controllable way, which is crucial for maintaining geometric consistency. 

Therefore, we propose a texture-geometry guided SDS (TG-SDS) loss within the latent space of ControlNet~\cite{zhang2023adding}. ControlNet allows for the integration of multi-conditional images to control image generation. Building on this capability, we propagate texture and geometric information from reference views to other views, using these as conditional images to guide ControlNet.

\begin{figure*}[!t]
    \centering    \includegraphics[width=\linewidth]{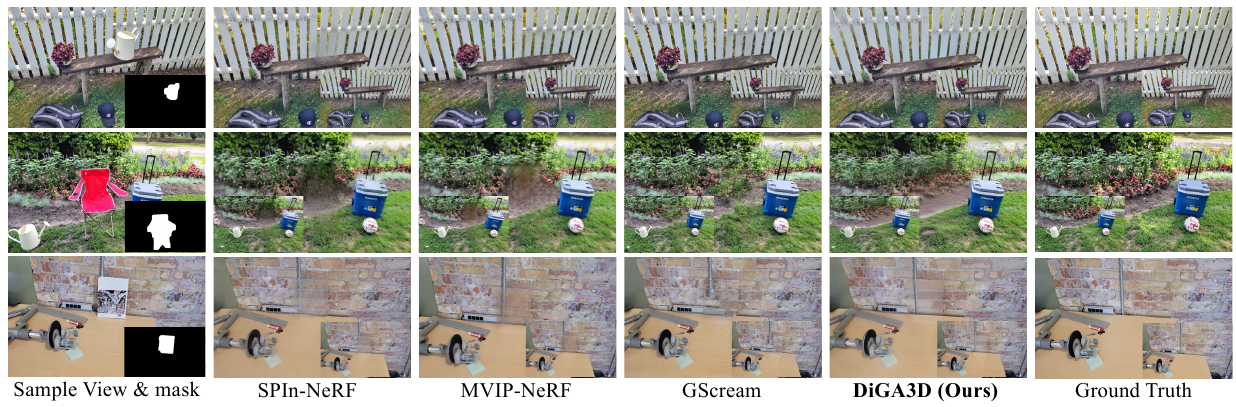}
    \vspace{-15pt}
    \caption{\small Qualitative results of the object removal task. For each scene, we present two novel views to compare the rendering quality and multi-view consistency with the existing state-of-the-art methods.}   \label{fig:object_removal}
    \vspace{-8pt}
\end{figure*}

\noindent\textbf{Texture-Geometry Warping.} We first employ the depth image-based rendering (DIBR) method~\cite{fehn2004depth} to warp images from the reference views to other views. As illustrated in the fine stage of Fig.~\ref{fig:method} (b), to mitigate errors caused by significant pose differences between views, the reference views from a given cluster are only warped to other views within the same cluster. The warping process within each cluster is conducted independently. Specifically, for a view $I_{i}$ in cluster $C_j$, we warp each pixel $q$ of the reference view $I_{R_j}$ within this cluster, along with its depth value $D_{R_j}$, estimated by a 2D depth estimator~\cite{ranftl2021vision} $\tilde{D}$. We then compute a wrapped pixel $q_{{R_j} \rightarrow i}$ as follows:
\begin{equation}\label{eq:warp}
    q_{{R_j} \rightarrow i} = \mathbf{K}\mathbf{P_i}\mathbf{P_{R_j}^{-1}}\mathbf{K^{-1}}[q, D_{R_j}],
\end{equation}
where $\mathbf{K}$, $\mathbf{P_i}$, $\mathbf{P_{R_j}}$ indicate the intrinsic matrix, the camera pose of view $i$, and the camera pose of reference view $R_j$, respectively. Through this process, we obtain the warped images $I'_i$ from reference views to other views. Additionally, we apply the Canny edge detector~\cite{canny1986computational} to generate texture maps $C'_i$ and employ a 2D depth estimator~\cite{ranftl2021vision} to produce the depth maps $D'_i$.


\begin{figure*}[!t]
    \centering    \includegraphics[width=0.95\linewidth]{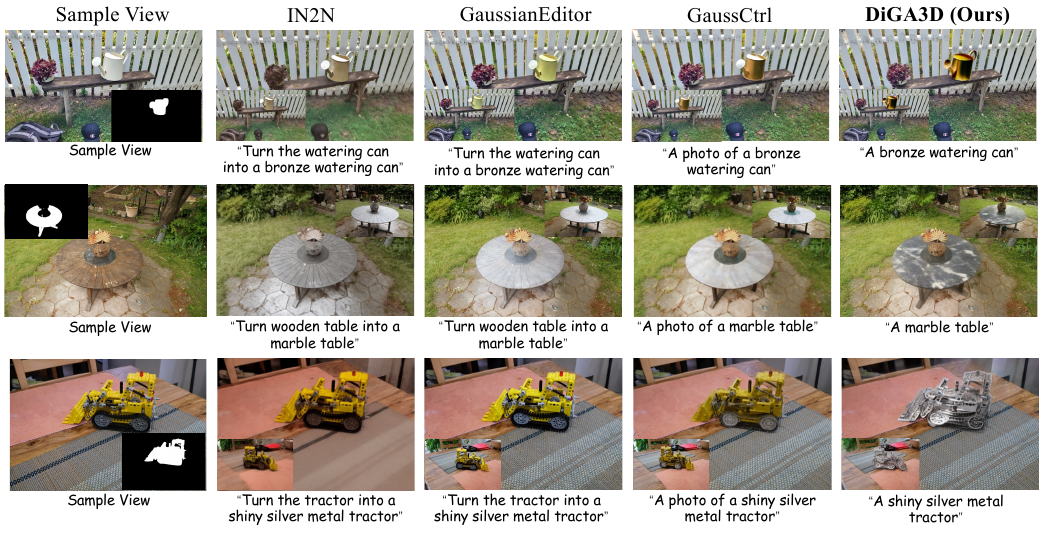}
    \vspace{-8pt}
    \caption{\small Qualitative results of the object re-texturing task. For each scene, we present two novel views to compare the rendering quality and multi-view consistency with the existing state-of-the-art methods.}
    \label{fig:object_editing}
    \vspace{-6pt}
\end{figure*}

\begin{table*}[t]
    \centering
    \setlength{\tabcolsep}{0.025\linewidth}
    \resizebox{\linewidth}{!}{
    \begin{tabular}{l|ccc|ccc}
    \toprule
         Methods& PSNR $\uparrow$ &SSIM $\uparrow$& LPIPS $\downarrow$& Masked PSNR$\uparrow$ & Masked SSIM$\uparrow$ & Masked LPIPS$\downarrow$ \\ \midrule
         SPIn-NeRF~\cite{mirzaei2023spin}& 20.32 &0.48 & 0.41 & 15.45 & 0.22& 0.56  \\
         NeRFiller~\cite{weber2024nerfiller}& 17.31 &0.28 & 0.43 & 15.48 & 0.24& 0.65 \\
         MVIP-NeRF~\cite{chen2024mvip}& 20.60 &0.49 & 0.44 & 10.00 & 0.24& 0.60 \\
         GScream~\cite{wang2024gscream}& 20.49 &\textbf{0.58} & \textbf{0.28} & 15.84 & 0.21& \textbf{0.54} \\
         \midrule
         \textbf{DiGA3D (Ours)}& \textbf{20.71} &\textbf{0.58} & \textbf{0.28} & \textbf{17.22} & \textbf{0.26}& 0.56  \\
    \bottomrule
    \end{tabular}}
    \vspace{-8pt}
    \caption{\small Quantitative results of the object removal task. We compared our method with four baselines, i.e., SPIn-NeRF~\cite{mirzaei2023spin}, NeRFiller~\cite{weber2024nerfiller}, MVIP-NeRF~\cite{chen2024mvip}, and GScream~\cite{wang2024gscream}. Our method achieves clear improvements in PSNR and obtains better scores in most metrics.}
    \label{tab:object_removal}
    \vspace{-15pt}
\end{table*}

\noindent\textbf{Multi-View SDS Loss.}
After acquiring conditional images with both texture and geometry details, \ie, the texture maps and depth maps derived from texture-geometry warping, we employ them to compute the TG-SDS loss in Fig.~\ref{fig:framework}. In this process, the rendered images $I_i$, along with the projected warped texture maps $C'$, warped depth maps $D'$, and mask $m$ are input into the multi-control diffusion model $\phi$ for conditioned generation: 
{\setlength\abovedisplayskip{2pt}
\setlength\belowdisplayskip{2pt}
\begin{equation}
\begin{split}
    \nabla_{\theta} \mathcal{L}_{\scalebox{0.5}{TG-SDS}} &= \mathbb{E}_{t,\epsilon}\left[ w(t) \big(\epsilon_\phi(I^i_t; m_i, y,t, \scalebox{0.9}{C}'_i, \scalebox{0.9}{D}'_i) - \epsilon^i\big)\frac{\partial I_i}{\partial \theta} \right],
\end{split}
\end{equation}}
where the noise latent $I^i_t$ is derived from the rendered images $I_i$ using the encoder of the diffusion model $\phi$, and $N$ is the numbers of rendered images. It is important to note that we only backpropagate the gradient for the masked pixels.

\subsection{Optimization}
\label{sec:optimization}
In the coarse stage, we employ a pre-trained monocular depth estimator~\cite{ranftl2021vision} $
\tilde{D}$ to produce the depth map $D_i$ from the inpainted image $I_i$. 
The 3D Gaussians $G$ are optimized with all properties by minimizing the photometric loss and depth loss:
{\setlength\abovedisplayskip{2pt}
\setlength\belowdisplayskip{2pt}
\begin{equation}
\begin{split}
    \mathcal{L}_{rgb} &= (1- \lambda)\mathcal{L}_1(\mathcal{R}(G)_I, I) \\
    &+ \lambda\mathcal{L}_{D-SSIM}(\mathcal{R}(G)_I, I),
\end{split}
\end{equation}}
where $\lambda$ = 0.2 is empirically set for all experiments. The depth loss can be represented as:
{\setlength\abovedisplayskip{2pt}
\setlength\belowdisplayskip{2pt}
\begin{align}\label{eq:loss_depth}
    \mathcal{L}_{depth} &= \mathcal{L}_1(\mathcal{R}(G)_D, D),
\end{align}}
Due to the monocular depth is not a metric depth, we align the monocular depth $D$ with the rendered depth $\mathcal{R}(G)_D$ using scale and shift parameters through least-squares estimation in Eq.~\ref{eq:warp} and Eq.~\ref{eq:loss_depth}.

In the fine stage, we refine the 3D Gaussians $G$ by optimizing with $\mathcal{L}_{TG-SDS}$. Consequently, the overall loss function is defined as:
\begin{equation}
    \mathcal{L} = \lambda_{rgb}\mathcal{L}_{rgb} + \lambda_{depth}\mathcal{L}_{depth} + \lambda_{\scalebox{0.5}{TG-SDS}}\mathcal{L}_{\scalebox{0.5}{TG-SDS}},
\end{equation}
 where $\lambda_{r}$, $\lambda_{d}$, and $\lambda_{\scalebox{0.5}{TG-SDS}}$ are the coefficients for photometric loss, depth loss, and TG-SDS loss, respectively.

%% file: sec/4_experiments.tex
\section{Experiment}

\begin{table}[t]
    \centering
    \setlength{\tabcolsep}{0.004\linewidth}
    \resizebox{0.9\linewidth}{!}{
    \begin{tabular}{l|cc|c}
    \toprule
         \multirow{2}{*}{Methods}& \multicolumn{2}{c|}{Re-texturing} & \multicolumn{1}{c}{Replacement}\\
         &$CLIP_{dir}$ $\uparrow$ & User Study(\%) & $CLIP_{dir}$ $\uparrow$ \\ \midrule
         IN2N~\cite{haque2023instruct}& 0.0572 &1.85\%& 0.0354  \\
         GaussianEditor~\cite{chen2024gaussianeditor}& 0.0702 &1.85\% & 0.0908 \\
         GaussCtrl~\cite{wu2024gaussctrl}& 0.0742 &12.97\% & 0.1097 \\ \midrule
         \textbf{DiGA3D (Ours)}& \textbf{0.1751} &\textbf{83.33\%} & \textbf{0.2247}  \\
    \bottomrule
    \end{tabular}}
    \vspace{-8pt}
    \caption{\small~Quantitative results of object re-texturing and replacement. We compared our method with three competitors, i.e., Instruct-NeRF2NeRF (IN2N)~\cite{haque2023instruct}, GaussianEditor~\cite{chen2024gaussianeditor}, and GaussCtrl~\cite{wu2024gaussctrl}.~{$CLIP_{dir}$:~CLIP Text-Image Direction Similarity.}}    \label{tab:editing_replace}
    \vspace{-18pt}
\end{table}


\begin{figure*}[t]
    \centering
    \includegraphics[width=0.97\linewidth]{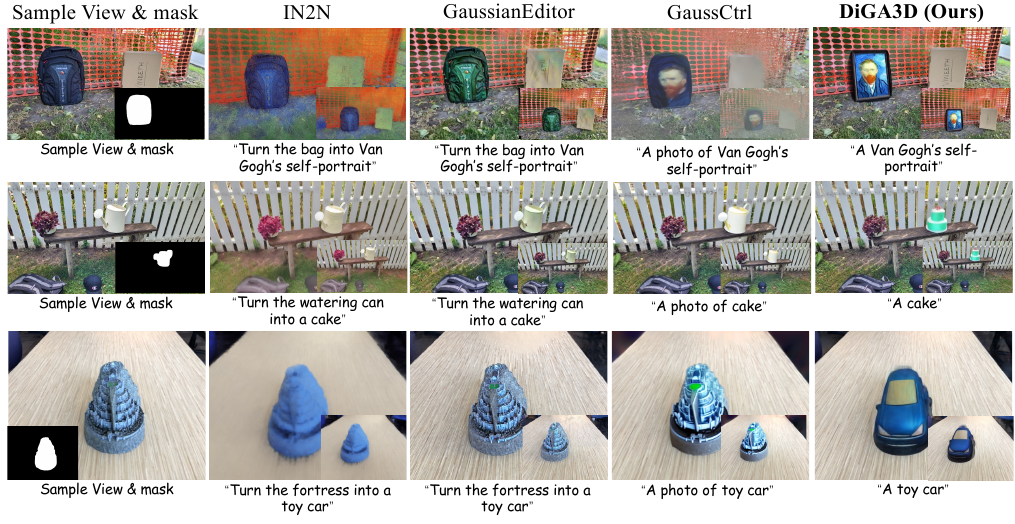}
    \vspace{-10pt}
    \caption{\small Qualitative results of the object replacement task. For each scene, we present two novel views to compare the rendering quality and multi-view consistency with the existing state-of-the-art methods.}
    \vspace{-15pt}
    \label{fig:object_replace}
\end{figure*}

\subsection{Experimental Setup}
\textbf{Datasets.}
We evaluate our versatile 3D inpainting methods in three different datasets with multi-view images from feed-forward and 360 degrees: 1) SPIn-NeRF dataset~\cite{mirzaei2023spin} provide 10 scenes that each scene includes 60 images with an unwanted object (training views) and 40 images without it (test views), it originally designed for object removal task but also can used for evaluating other inpainting tasks. 2) MipNeRF360~\cite{barron2022mip} dataset. 3) LLFF dataset~\cite{mildenhall2019llff}.

\noindent\textbf{Evaluation Metrics.}
To evaluate the effectiveness of our method for versatile 3D inpainting, we employ different evaluation metrics tailored to specific tasks. 1) For the object removal task, we evaluate our method using PSNR, SSIM, and LPIPS scores on the SPIn-NeRF dataset~\cite{mirzaei2023spin}. 2) For object re-texturing and replacement tasks, we follow established practices by calculating the CLIP score and conducting a user study to measure the fidelity between our method and previous approaches. 
Specifically, we utilize CLIP Text Image Directional Similarity ($CLIP_{dir}$) to assess how well object re-texturing and replacement align with text instructions. Additionally, the user study includes a four-way voting process to evaluate and compare our method with other state-of-the-art methods in object re-texturing tasks.

\noindent\textbf{Implementation Details.}
Our method is implemented using the PyTorch library~\citep{paszke2017automatic}. For the coarse stage, we use PowerPaint-v1~\cite{zhuang2023task} and Stable Diffusion v1.5 with its ControlNet~\cite{zhang2023adding} as our 2D inpainter. We employ Stable Diffusion v1.5 and its corresponding ControlNet from the Hugging Face library to guide our TG-SDS loss. To generate 2D masks for inpainting, we utilize Lang SAM~\citep{kirillov2023segment} based on mask prompts. In addition, we build upon Scaffold-GS~\cite{lu2024scaffold} for our 3D representations. Our method is trained on a single NVIDIA 48GB A6000 GPU.
\subsection{Methods for Comparison}
To assess our methods across various inpainting tasks, we conduct comparisons with different techniques tailored for each specific task. For object removal, we compare our approach with SPIn-NeRF~\cite{mirzaei2023spin}, NeRFiller~\cite{weber2024nerfiller}, MVIP-NeRF~\cite{chen2024mvip}, and GScream~\cite{wang2024gscream}. For object re-texturing and replacement, we evaluate our method against Instruct-NeRF2NeRF (IN2N)~\cite{haque2023instruct}, GaussianEditor~\cite{chen2024gaussianeditor}, and GaussCtrl~\cite{wu2024gaussctrl}.
\subsection{Results}
We primarily provide quantitative and qualitative comparisons of three inpainting tasks, \ie, object removal, object re-texturing, and object replacement, to evaluate the effectiveness of our versatile 3D inpainting framework.
\subsubsection{Object Removal}
Quantitative and qualitative comparisons between our method and three baseline methods are illustrated in Tab.~\ref{tab:object_removal} and Fig.~\ref{fig:object_removal}, respectively.

\noindent\textbf{Quantitative Comparison.}
As demonstrated in Tab.~\ref{tab:object_removal}, our methods outperform or match SPIn-NeRF, NeRFiller, and MVIP-NeRF across all evaluated metrics. While our rendering results exhibit some limitations in the masked LPIPS compared to GScream, we achieve a comparable score in this metric and show significant advantages in PSNR, masked PSNR, and masked SSIM when compared to all other methods. These results indicate the effectiveness of our approach.

\noindent\textbf{Qualitative Comparison.} 
Fig.~\ref{fig:object_removal} presents qualitative results across three scenes from the SPIn-NeRF dataset. The leftmost column displays randomly selected scene images along with their corresponding masks. In the first scene, our method shows minimal artifacts in the removal areas, which are especially evident in the results of the first row. In the second and third scenes, our approach consistently achieves cross-view and contextual coherence. Compared to SPIn-NeRF and MVIP-NeRF, our method captures more details with fewer artifacts, showcasing our better ability.


\subsubsection{Object Re-Texturing}
\noindent\textbf{Quantitative Comparison.}
Tab. \ref{tab:editing_replace} presents the $CLIP_{dir}$ scores and the results of the user study. For the $CLIP_{dir}$ scores, we averaged the scores across six scenes from the SPIn-NeRF~\cite{mirzaei2023spin} and MipNeRF360~\cite{barron2022mip} datasets. Based on the $CLIP_{dir}$ score, our methods show significant advantages over other approaches, indicating a higher alignment of our re-texturing results with various text instructions. Additionally, we conducted a user study employing a four-way voting process, allowing users to select the most relevant edited scene based on the text prompts while ensuring high rendering quality. The results indicate that our methods also exhibit advantages compared to other methods.

\noindent\textbf{Qualitative Comparison.}
Fig.~\ref{fig:object_editing} showcases diverse re-texturing results of our method in both forward-facing and 360-degree scenes. We leverage different text instructions to assess our approach and compare it with three previous works. The qualitative results demonstrate that our method aligns more closely with the text prompts. 
\subsubsection{Object Replacement}
\noindent\textbf{Quantitative Comparison.}
The quantitative comparison results for object replacement are presented in Tab. \ref{tab:editing_replace}. We find that our methods achieve relatively high scores compared to other approaches, demonstrating that they can generate more realistic and relevant objects with text prompts.

\noindent\textbf{Qualitative Comparison.}
We present qualitative comparison results in Fig.~\ref{fig:object_replace}. It is evident that previous methods can only generate objects with similar styles based on text prompts and struggle to implement significant geometric changes, whereas our approach can replace objects and seamlessly complete regions with contextual consistency.

\begin{figure}[!t]
    \centering
    \includegraphics[width=\linewidth]{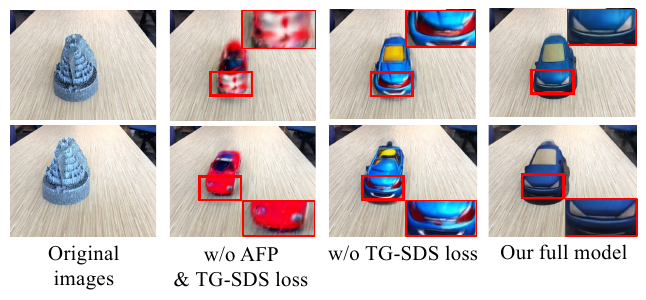}
    \vspace{-20pt}
    \caption{The visualization of ablation study for key components on the object replacement task using LLFF dataset~\cite{mildenhall2019llff}.}
    \label{fig:ablation}
    \vspace{-8pt}
\end{figure}

\begin{table}[t]
    \centering
    \setlength{\tabcolsep}{0.004\linewidth}
    \resizebox{0.78\linewidth}{!}{
    \begin{tabular}{l|ccc}
    \toprule
         Methods& PSNR $\uparrow$ &SSIM $\uparrow$& LPIPS$\downarrow$ \\ \midrule
         Our full model & \textbf{20.71} &\textbf{0.58} & \textbf{0.28}\\ 
         w/o TG-SDS loss& 20.66 &0.57 & 0.29 \\
         w/o AFP \& TG-SDS loss & 20.45 &0.57 & 0.30 \\       
    \bottomrule
    \end{tabular}}
    \vspace{-8pt}
    \caption{\small The quantitative ablation study of key components on the object removal task using SPIn-NeRF dataset~\cite{mirzaei2023spin}.}
    \label{tab:ablation}
    \vspace{-10pt}
\end{table}

\subsection{Ablation Study}

We conduct ablation experiments on our key components, reference view selection, and TG-SDS loss.

\noindent\textbf{Quantitative Analysis of Key Components.} As detailed in Tab.~\ref{tab:ablation}, we gradually assess our baseline (w/o AFP \& TG-SDS loss), coarse stage (w/o TG-SDS loss), and our fine stage (full model). In the baseline, we solely utilize the 2D inpainter~\cite{zhuang2023task, zhang2023adding} and depend on the convergence of 3D representations. By integrating DDIM inversion and AFP within the 2D inpainter, we achieve a notable 0.21 improvement in PSNR, indicating significant enhancements. With the addition of our fine stage, all three metrics exhibit further improvements, underscoring the effectiveness of key component of our method.

\noindent\textbf{Qualitative Analysis of Key Components.} In Fig.~\ref{fig:ablation}, we depict the visualizations of the ablation study on key components. We provide an example of replacing the `fortress' with `a toy car'. Starting with our baseline in the second column, noticeable blurriness is observed within the inpainting regions, stemming from the inconsistencies in the 2D inpainter's direct inpainting results. By employing AFP, we have significantly improved the issue of inconsistencies, although some artifacts and texture details still lack consistency. With the addition of the fine stage, our full model exhibits more consistent and smoother appearance results.

\begin{figure}[!t]
    \centering
    \includegraphics[width=\linewidth]{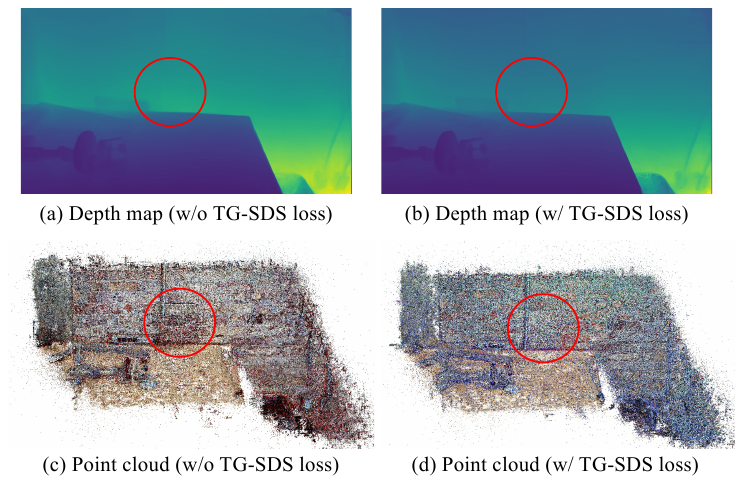}
    \vspace{-22pt}
    \caption{Qualitative ablation study for the proposed TG-SDS optimization loss on the SPIn-NeRF dataset~\cite{mirzaei2023spin}.}
    \label{fig:ablation_geo}
    \vspace{-8pt}
\end{figure}

\begin{table}[t]
    \centering
    \resizebox{\linewidth}{!}{
    \begin{tabular}{c|ccccc}
    \toprule
        $K$ & PSNR$\uparrow$ & SSIM$\uparrow$  &LPIPS$\downarrow$  & Memory & Resolution \\ \midrule
         1&19.87&0.4670&0.3350& 41G& $512 \times 904$\\ 
         2&19.89 &0.4665&0.3412& 46G &$512 \times 904$\\ 
         3&\textbf{19.94} &\textbf{0.4676}&\textbf{0.3330}& 47G &$512 \times 904$\\ 
    \bottomrule
    \end{tabular}}
    \vspace{-8pt}
    \caption{The selection of hyperparameter K. We evaluate different values of K on Scene 1 of the SPIn-NeRF~\cite{mirzaei2023spin} dataset using a single A6000 GPU.}
    \label{tab:k}
    \vspace{-13pt}
\end{table}


\noindent\textbf{Quantitative Analysis of Hyperparameter K in Reference View Selection.}
When using K-means for selecting reference views, it is important to balance memory cost and performance during the coarse stage. As demonstrated in Tab.~\ref{tab:k}, we achieve this balance by choosing $K=3$ for our experiments on the SPIn-NeRF~\cite{mirzaei2023spin}, which ensures both high performance and the ability to run efficiently on a single A6000 GPU.

\noindent\textbf{Qualitative Analysis of TG-SDS Loss.} Due to the monocular depth supervision in the coarse stage, we further conduct an ablation study to analyze the role of TG-SDS loss in geometric regularization. In Fig. \ref{fig:ablation_geo} (a) and (b), both the depth maps without TG-SDS loss and with TG-SDS loss are free from artifacts related to foreground objects, with minimal differences between them. Subsequently, we visualize the point clouds for both cases in Fig. \ref{fig:ablation_geo} (c) and (d), respectively. It is evident that (c) clearly exhibits remnants of foreground objects and some redundant points, whereas (d) showcases improved geometries and textures, demonstrating the effectiveness of our TG-SDS loss in enhancing geometry.

%% file: sec/5_conclusion.tex
\section{Conclusion}
In this paper, we introduce a versatile 3D inpainting pipeline that leverages diffusion models to propagate consistent appearance and geometry using a coarse-to-fine strategy. Specifically, we utilize K-means clustering to select reference views that capture the majority of appearance and geometry information across the entire scene. During the coarse stage, we perform multi-view inpainting by propagating attention features from reference views to other views through the latent space of a 2D inpainter. In the fine stage, we introduce the TG-SDS loss to further regularize the geometry of the inpainted 3D scene. We conduct extensive experiments on multiple 3D inpainting tasks to demonstrate the effectiveness of our method.

\noindent\textbf{Limitations and Future Work:} The object replacement task in 360-degree scenes may encounter multi-face Janus problems when the replaced object significantly differs in shape and appearance from different views. We aim to address this issue by designing view priors in the future.


%% file: sec/6_suppl.tex

\section{Appendix}
\subsection{Additional Implementation Details}
\label{sec:implementation}
To select reference views, we utilize K-means clustering with $K = 3$ to identify the views that are closest to the cluster centers as our reference views. During the coarse stage, we set $\lambda = 0.6$ for our AFP mechanism to propagate reference attention features into other attention features effectively. In the fine stage, we set the guidance scale to 7.5, the condition scale for depth to 1.0, and the condition scale for texture to 0.8. Some parameters will be adjusted based on the specific scenario. 
  
\noindent\textbf{Discussion on K-means for selecting reference views.} We compared the method of selecting reference views using K-means clustering with the method of randomly selecting reference views on the object removal task using the ground truth SPIn-NeRF dataset~\cite{mirzaei2023spin}. We found that in the coarse stage, there was not much difference between the two methods in propagating attention features from reference views to other views. However, in the fine stage, the reference views selected by K-means produced more stable clusters, resulting in more consistent and accurate outcomes when warping reference views to other views.

\subsection{Ablations on Using Different 2D Inpainters}
\begin{figure}[t]
    \centering
    \includegraphics[width=\linewidth]{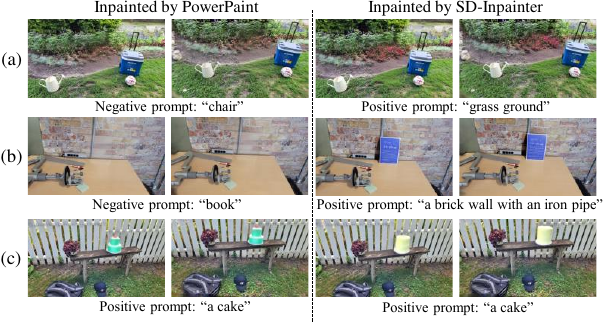}
    \vspace{-10pt}
    \caption{Ablations on using different 2D inpainters, \ie, PowerPaint~\cite{zhuang2023task} and SD-Inpainter~\cite{rombach2022high}. (a) and (b) display comparisons for object removal tasks, whereas (c) presents comparisons for object replacement tasks.}
    \label{fig:suppl_inpainter}
    \vspace{-10pt}
\end{figure}

We conduct qualitative ablation studies using different text-guided 2D inpainters, specifically SD-Inpainter~\cite{rombach2022high} and PowerPaint~\cite{zhuang2023task}, within our methods applied to the SPIn-NeRF~\cite{mirzaei2023spin} datasets. As shown in Fig.~\ref{fig:suppl_inpainter}, our method achieves consistent inpainting results across different 2D inpainters. We observe in (b) that the SD-Inpainter sometimes struggles to deliver successful removal results with complex prompts. In contrast, PowerPaint effectively uses negative prompts to describe the objects to be removed, yielding more accurate results.

\subsection{Ablations on TG-SDS loss}
\begin{figure}[t]
    \centering
    \includegraphics[width=\linewidth]{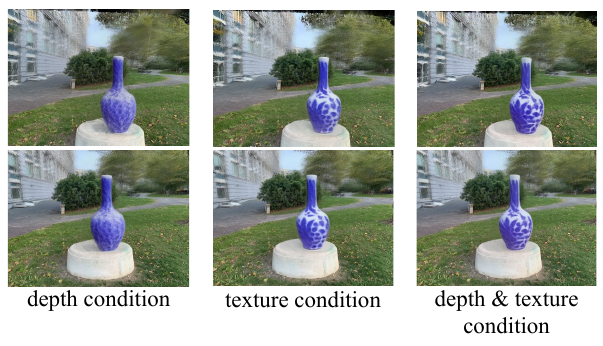}
    \vspace{-20pt}
    \caption{Additional ablation study on the TG-SDS loss.}
    \label{fig:suppl_sds}
    \vspace{-10pt}
\end{figure}

As illustrated in Fig.~\ref{fig:suppl_sds}, we conduct additional ablation studies on our TG-SDS loss with positive text prompts, such as `a vase textured with some flowers', which includes intricate texture details and specific geometry. By integrating both texture and depth conditions into the TG-SDS loss, we can achieve improved texture and detailed geometry not only for foreground objects but also for the background.

\subsection{Additional Quantitative Results}
We present additional no-reference measurements on two key metrics, specifically MUSIQ~\cite{ke2021musiq} and Corrs (number of high-quality correspondences between random pairs of frames). These metrics are commonly utilized to evaluate the aesthetic and geometric quality of images. We provide a comparison with NeRFiller across various scenes, demonstrating the capability of both object removal and replacement tasks. As indicated in Tab.~\ref{tab:suppl_no-reference}, our method achieves significantly superior results on both MUSIQ and Corrs metrics, underscoring the enhanced aesthetic and geometric quality facilitated by our approach.

\begin{table}[h]
    \centering
    \setlength{\tabcolsep}{0.03\linewidth}
    \resizebox{\linewidth}{!}{
        \begin{tabular}{l|cc|cc}
            \toprule
            \multirow{2}{*}{Methods} & \multicolumn{2}{c|}{Removal} & \multicolumn{2}{c}{Replacement}\\
            & MUSIQ $\uparrow$ &Corrs $\uparrow$ & MUSIQ $\uparrow$ &Corrs $\uparrow$\\ \midrule
            NeRFiller~\cite{weber2024nerfiller}& 65.55 &7343  &65.25&7223\\
            \textbf{DiGA3D (Ours)}& \textbf{68.89}&\textbf{7421}&\textbf{68.70}&\textbf{7512}  \\
            \bottomrule
    \end{tabular}}
    \vspace{-10pt}
    \caption{\small Results of the two tasks with MUSIQ and Corrs.}
    \label{tab:suppl_no-reference}
    \vspace{-15pt}
\end{table}

\subsection{Additional Qualitative Results}
We provide supplementary qualitative results for a range of inpainting tasks utilizing the SPIn-NeRF dataset~\cite{mirzaei2023spin}, LLFF dataset~\cite{mildenhall2019llff}, MipNeRF360 dataset~\cite{barron2022mip}, and Instruct-NeRF2NeRF dataset~\cite{haque2023instruct}. 

\begin{figure}[t]
    \centering
    \includegraphics[width=\linewidth]{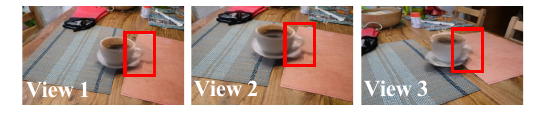}
    \vspace{-20pt}
    \caption{A failure case of the object replacement task.}
    \label{fig:suppl_failure_case}
    \vspace{-10pt}
\end{figure}

\subsubsection{Additional Results for Object Removal}
\begin{figure*}[t]
    \centering
    \includegraphics[width=\linewidth]{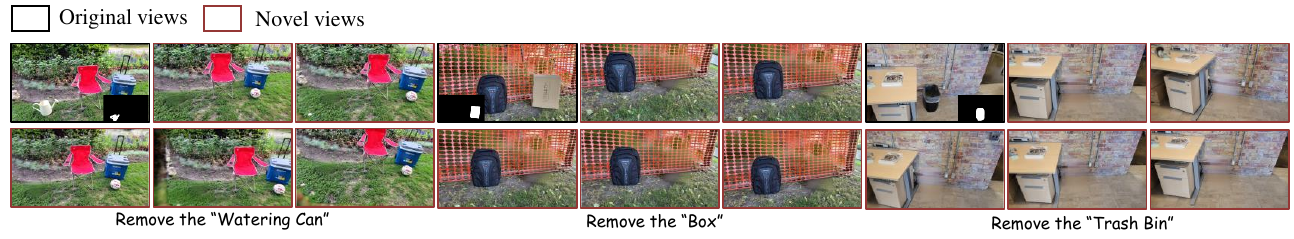}
    \vspace{-20pt}
    \caption{Additional object removal results.}
    \vspace{-10pt}
    \label{fig:suppl_object_removal}
\end{figure*}
As presented in Fig.~\ref{fig:suppl_object_removal}, we present three additional object removal examples across different scenes from the SPIn-NeRF~\cite{mirzaei2023spin} dataset. In the first two scenes, we successfully remove objects that lack corresponding ground truth data in the original dataset. This removal is achieved using text prompts.

\subsubsection{Additional Results for Object Re-Texturing}
\begin{figure*}[t]
    \centering
    \includegraphics[width=0.9\linewidth]{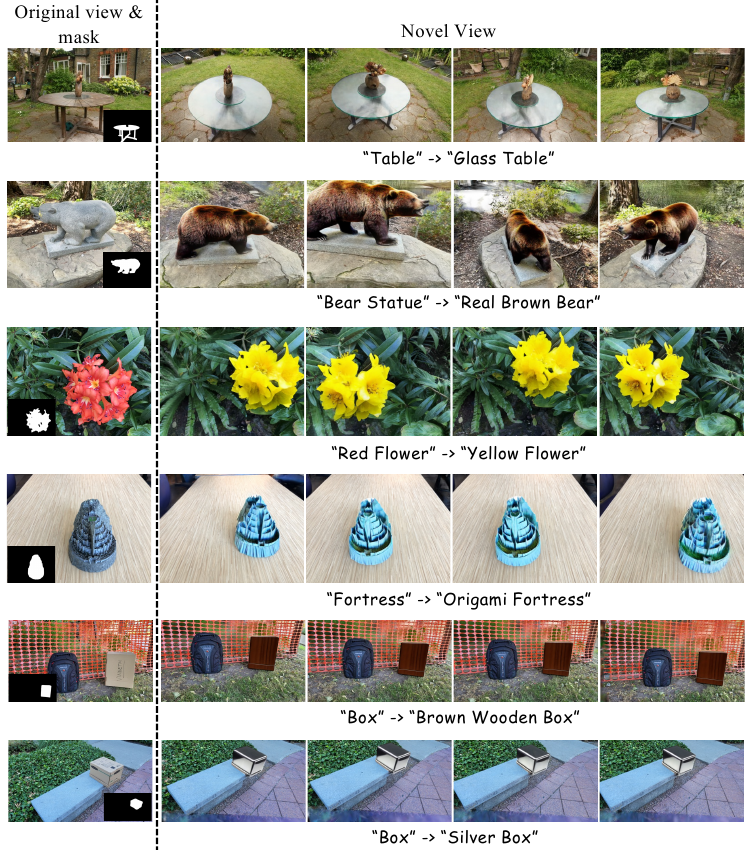}
    \caption{Additional object re-texturing results.}
    \label{fig:suppl_object_editing}
\end{figure*}

In Fig.~\ref{fig:suppl_object_editing}, we present additional object re-texturing results across various scenes and prompts. These further demonstrate the effectiveness of our method.

\subsubsection{Additional Results for Object Replacement}
\begin{figure*}[t]
    \centering
    \includegraphics[width=0.9\linewidth]{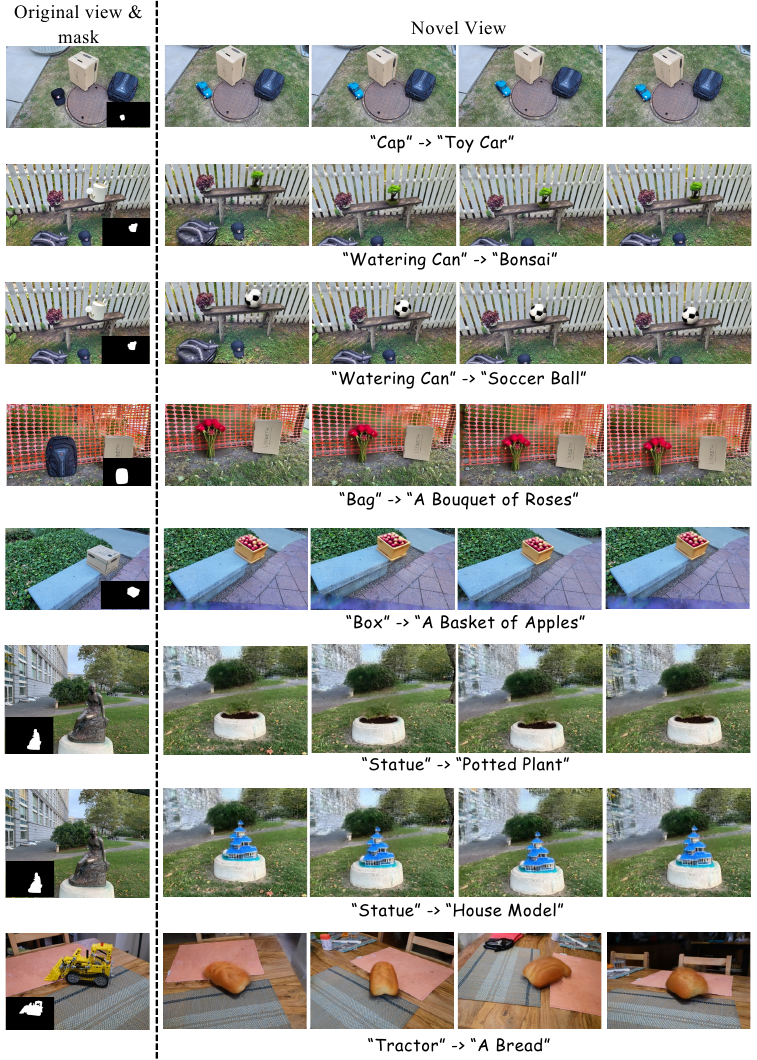}
    \caption{Additional object replacement results.}
    \label{fig:suppl_object_replace}
\end{figure*}

Furthermore, as illustrated in Fig.~\ref{fig:suppl_object_replace}, we present additional object replacement results to further evaluate the diversity and generalizability of our methods. By employing different text prompts within a single scene, we produce various object replacement outcomes. 

\subsection{Details of User Study}
Similar to GaussianEditor~\cite{chen2024gaussianeditor}, we created six questions with the videos of novel view rendering results for the object re-texturing task questionnaire (including the scenes presented in our main paper), each featuring the original scene, text instructions, and re-texturing results from IN2N~\cite{haque2023instruct}, GaussianEditor~\cite{chen2024gaussianeditor}, GaussCtrl~\cite{wu2024gaussctrl}, and our method, all labeled randomly. Participants selected their preferred outcome, and after 18 participants completed the questionnaires, we collected a total of 108 votes.

\subsection{Analysis of Failure Cases}
As shown in Figure~\ref{fig:suppl_failure_case}, we show a failure case where we attempt to `replace the tractor with a cup of coffee'. The handle of the coffee cup is visible in multiple views, causing a multi-face issue. This common challenge may arise from the substantial geometric changes from a tractor to a coffee cup, and the limitation of the diffusion model and SDS optimization for fine-grained geometric inpainting, particularly noticeable in view 3.

%% file: sec/7_ack.tex
\section{Acknowledgments}
The work of Qiong Luo and Jingyi Pan is supported by a startup grant from the Hong Kong University of Science and Technology (Guangzhou). The work of Dan Xu is supported in part by the Early Career Scheme of the Research Grants Council (RGC) of the Hong Kong SAR under grant No. 26202321, SAIL Research Project, and HKUST-Zeekr Collaborative Research Fund.

%% file: a_arxiv_version.bbl
\begin{thebibliography}{49}
\providecommand{\natexlab}[1]{#1}
\providecommand{\url}[1]{\texttt{#1}}
\expandafter\ifx\csname urlstyle\endcsname\relax
  \providecommand{\doi}[1]{doi: #1}\else
  \providecommand{\doi}{doi: \begingroup \urlstyle{rm}\Url}\fi

\bibitem[Barron et~al.(2022)Barron, Mildenhall, Verbin, Srinivasan, and Hedman]{barron2022mip}
Jonathan~T Barron, Ben Mildenhall, Dor Verbin, Pratul~P Srinivasan, and Peter Hedman.
\newblock Mip-nerf 360: Unbounded anti-aliased neural radiance fields.
\newblock In \emph{Proceedings of the IEEE/CVF conference on computer vision and pattern recognition}, pages 5470--5479, 2022.

\bibitem[Bartrum et~al.(2024)Bartrum, Nguyen-Phuoc, Xie, Li, Khan, Avetisyan, Lanman, and Xiao]{bartrum2024replaceanything3d}
Edward Bartrum, Thu Nguyen-Phuoc, Chris Xie, Zhengqin Li, Numair Khan, Armen Avetisyan, Douglas Lanman, and Lei Xiao.
\newblock Replaceanything3d: Text-guided 3d scene editing with compositional neural radiance fields.
\newblock \emph{arXiv preprint arXiv:2401.17895}, 2024.

\bibitem[Canny(1986)]{canny1986computational}
John Canny.
\newblock A computational approach to edge detection.
\newblock \emph{IEEE Transactions on pattern analysis and machine intelligence}, \penalty0 (6):\penalty0 679--698, 1986.

\bibitem[Cao et~al.(2024{\natexlab{a}})Cao, Cai, Dong, Wang, and Fu]{cao2024leftrefill}
Chenjie Cao, Yunuo Cai, Qiaole Dong, Yikai Wang, and Yanwei Fu.
\newblock Leftrefill: Filling right canvas based on left reference through generalized text-to-image diffusion model.
\newblock In \emph{Proceedings of the IEEE/CVF Conference on Computer Vision and Pattern Recognition}, pages 7705--7715, 2024{\natexlab{a}}.

\bibitem[Cao et~al.(2024{\natexlab{b}})Cao, Yu, Fu, Wang, and Xue]{cao2024mvinpainter}
Chenjie Cao, Chaohui Yu, Yanwei Fu, Fan Wang, and Xiangyang Xue.
\newblock Mvinpainter: Learning multi-view consistent inpainting to bridge 2d and 3d editing.
\newblock \emph{arXiv preprint arXiv:2408.08000}, 2024{\natexlab{b}}.

\bibitem[Caron et~al.(2021)Caron, Touvron, Misra, J{\'e}gou, Mairal, Bojanowski, and Joulin]{caron2021emerging}
Mathilde Caron, Hugo Touvron, Ishan Misra, Herv{\'e} J{\'e}gou, Julien Mairal, Piotr Bojanowski, and Armand Joulin.
\newblock Emerging properties in self-supervised vision transformers.
\newblock In \emph{Proceedings of the IEEE/CVF international conference on computer vision}, pages 9650--9660, 2021.

\bibitem[Chen et~al.(2024{\natexlab{a}})Chen, Loy, and Pan]{chen2024mvip}
Honghua Chen, Chen~Change Loy, and Xingang Pan.
\newblock Mvip-nerf: Multi-view 3d inpainting on nerf scenes via diffusion prior.
\newblock In \emph{Proceedings of the IEEE/CVF Conference on Computer Vision and Pattern Recognition}, pages 5344--5353, 2024{\natexlab{a}}.

\bibitem[Chen et~al.(2024{\natexlab{b}})Chen, Chen, Zhang, Wang, Yang, Wang, Cai, Yang, Liu, and Lin]{chen2024gaussianeditor}
Yiwen Chen, Zilong Chen, Chi Zhang, Feng Wang, Xiaofeng Yang, Yikai Wang, Zhongang Cai, Lei Yang, Huaping Liu, and Guosheng Lin.
\newblock Gaussianeditor: Swift and controllable 3d editing with gaussian splatting.
\newblock In \emph{Proceedings of the IEEE/CVF Conference on Computer Vision and Pattern Recognition}, pages 21476--21485, 2024{\natexlab{b}}.

\bibitem[Efros and Leung(1999)]{efros1999texture}
Alexei~A Efros and Thomas~K Leung.
\newblock Texture synthesis by non-parametric sampling.
\newblock In \emph{Proceedings of the seventh IEEE international conference on computer vision}, pages 1033--1038. IEEE, 1999.

\bibitem[Fehn(2004)]{fehn2004depth}
Christoph Fehn.
\newblock Depth-image-based rendering (dibr), compression, and transmission for a new approach on 3d-tv.
\newblock In \emph{Stereoscopic displays and virtual reality systems XI}, pages 93--104. SPIE, 2004.

\bibitem[Gao et~al.(2020)Gao, Saraf, Huang, and Kopf]{gao2020flow}
Chen Gao, Ayush Saraf, Jia-Bin Huang, and Johannes Kopf.
\newblock Flow-edge guided video completion.
\newblock In \emph{Computer Vision--ECCV 2020: 16th European Conference, Glasgow, UK, August 23--28, 2020, Proceedings, Part XII 16}, pages 713--729. Springer, 2020.

\bibitem[Haque et~al.(2023)Haque, Tancik, Efros, Holynski, and Kanazawa]{haque2023instruct}
Ayaan Haque, Matthew Tancik, Alexei~A Efros, Aleksander Holynski, and Angjoo Kanazawa.
\newblock Instruct-nerf2nerf: Editing 3d scenes with instructions.
\newblock In \emph{Proceedings of the IEEE/CVF International Conference on Computer Vision}, pages 19740--19750, 2023.

\bibitem[Ke et~al.(2021)Ke, Wang, Wang, Milanfar, and Yang]{ke2021musiq}
Junjie Ke, Qifei Wang, Yilin Wang, Peyman Milanfar, and Feng Yang.
\newblock Musiq: Multi-scale image quality transformer.
\newblock In \emph{Proceedings of the IEEE/CVF international conference on computer vision}, pages 5148--5157, 2021.

\bibitem[Kerbl et~al.(2023)Kerbl, Kopanas, Leimk{\"u}hler, and Drettakis]{kerbl20233d}
Bernhard Kerbl, Georgios Kopanas, Thomas Leimk{\"u}hler, and George Drettakis.
\newblock 3d gaussian splatting for real-time radiance field rendering.
\newblock \emph{ACM Trans. Graph.}, 42\penalty0 (4):\penalty0 139--1, 2023.

\bibitem[Kirillov et~al.(2023)Kirillov, Mintun, Ravi, Mao, Rolland, Gustafson, Xiao, Whitehead, Berg, Lo, et~al.]{kirillov2023segment}
Alexander Kirillov, Eric Mintun, Nikhila Ravi, Hanzi Mao, Chloe Rolland, Laura Gustafson, Tete Xiao, Spencer Whitehead, Alexander~C Berg, Wan-Yen Lo, et~al.
\newblock Segment anything.
\newblock In \emph{Proceedings of the IEEE/CVF International Conference on Computer Vision}, pages 4015--4026, 2023.

\bibitem[Li et~al.(2022)Li, Lu, Qin, Guo, and Cheng]{li2022towards}
Zhen Li, Cheng-Ze Lu, Jianhua Qin, Chun-Le Guo, and Ming-Ming Cheng.
\newblock Towards an end-to-end framework for flow-guided video inpainting.
\newblock In \emph{Proceedings of the IEEE/CVF conference on computer vision and pattern recognition}, pages 17562--17571, 2022.

\bibitem[Liu et~al.(2022)Liu, Shen, Chen, et~al.]{liu2022nerf}
Hao-Kang Liu, I Shen, Bing-Yu Chen, et~al.
\newblock Nerf-in: Free-form nerf inpainting with rgb-d priors.
\newblock \emph{arXiv preprint arXiv:2206.04901}, 2022.

\bibitem[Liu et~al.(2024{\natexlab{a}})Liu, Zhang, Li, Lin, and Jia]{liu2024video}
Shaoteng Liu, Yuechen Zhang, Wenbo Li, Zhe Lin, and Jiaya Jia.
\newblock Video-p2p: Video editing with cross-attention control.
\newblock In \emph{Proceedings of the IEEE/CVF Conference on Computer Vision and Pattern Recognition}, pages 8599--8608, 2024{\natexlab{a}}.

\bibitem[Liu et~al.(2024{\natexlab{b}})Liu, Ouyang, Wang, Cheng, Xiao, Zhu, Xue, Liu, Shen, and Cao]{liu2024infusion}
Zhiheng Liu, Hao Ouyang, Qiuyu Wang, Ka~Leong Cheng, Jie Xiao, Kai Zhu, Nan Xue, Yu Liu, Yujun Shen, and Yang Cao.
\newblock Infusion: Inpainting 3d gaussians via learning depth completion from diffusion prior.
\newblock \emph{arXiv preprint arXiv:2404.11613}, 2024{\natexlab{b}}.

\bibitem[Lloyd(1982)]{lloyd1982least}
Stuart Lloyd.
\newblock Least squares quantization in pcm.
\newblock \emph{IEEE transactions on information theory}, 28\penalty0 (2):\penalty0 129--137, 1982.

\bibitem[Lu et~al.(2024{\natexlab{a}})Lu, Yu, Xu, Xiangli, Wang, Lin, and Dai]{lu2024scaffold}
Tao Lu, Mulin Yu, Linning Xu, Yuanbo Xiangli, Limin Wang, Dahua Lin, and Bo Dai.
\newblock Scaffold-gs: Structured 3d gaussians for view-adaptive rendering.
\newblock In \emph{Proceedings of the IEEE/CVF Conference on Computer Vision and Pattern Recognition}, pages 20654--20664, 2024{\natexlab{a}}.

\bibitem[Lu et~al.(2024{\natexlab{b}})Lu, Ma, and Yin]{lu2024view}
Yiren Lu, Jing Ma, and Yu Yin.
\newblock View-consistent object removal in radiance fields.
\newblock In \emph{Proceedings of the 32nd ACM International Conference on Multimedia}, pages 3597--3606, 2024{\natexlab{b}}.

\bibitem[Mildenhall et~al.(2019)Mildenhall, Srinivasan, Ortiz-Cayon, Kalantari, Ramamoorthi, Ng, and Kar]{mildenhall2019llff}
Ben Mildenhall, Pratul~P. Srinivasan, Rodrigo Ortiz-Cayon, Nima~Khademi Kalantari, Ravi Ramamoorthi, Ren Ng, and Abhishek Kar.
\newblock Local light field fusion: Practical view synthesis with prescriptive sampling guidelines.
\newblock \emph{ACM Transactions on Graphics (TOG)}, 2019.

\bibitem[Mildenhall et~al.(2021)Mildenhall, Srinivasan, Tancik, Barron, Ramamoorthi, and Ng]{mildenhall2021nerf}
Ben Mildenhall, Pratul~P Srinivasan, Matthew Tancik, Jonathan~T Barron, Ravi Ramamoorthi, and Ren Ng.
\newblock Nerf: Representing scenes as neural radiance fields for view synthesis.
\newblock \emph{Communications of the ACM}, 65\penalty0 (1):\penalty0 99--106, 2021.

\bibitem[Mirzaei et~al.(2023{\natexlab{a}})Mirzaei, Aumentado-Armstrong, Brubaker, Kelly, Levinshtein, Derpanis, and Gilitschenski]{mirzaei2023reference}
Ashkan Mirzaei, Tristan Aumentado-Armstrong, Marcus~A Brubaker, Jonathan Kelly, Alex Levinshtein, Konstantinos~G Derpanis, and Igor Gilitschenski.
\newblock Reference-guided controllable inpainting of neural radiance fields.
\newblock In \emph{Proceedings of the IEEE/CVF international conference on computer vision}, pages 17815--17825, 2023{\natexlab{a}}.

\bibitem[Mirzaei et~al.(2023{\natexlab{b}})Mirzaei, Aumentado-Armstrong, Derpanis, Kelly, Brubaker, Gilitschenski, and Levinshtein]{mirzaei2023spin}
Ashkan Mirzaei, Tristan Aumentado-Armstrong, Konstantinos~G Derpanis, Jonathan Kelly, Marcus~A Brubaker, Igor Gilitschenski, and Alex Levinshtein.
\newblock Spin-nerf: Multiview segmentation and perceptual inpainting with neural radiance fields.
\newblock In \emph{Proceedings of the IEEE/CVF Conference on Computer Vision and Pattern Recognition}, pages 20669--20679, 2023{\natexlab{b}}.

\bibitem[Paszke et~al.(2017)Paszke, Gross, Chintala, Chanan, Yang, DeVito, Lin, Desmaison, Antiga, and Lerer]{paszke2017automatic}
Adam Paszke, Sam Gross, Soumith Chintala, Gregory Chanan, Edward Yang, Zachary DeVito, Zeming Lin, Alban Desmaison, Luca Antiga, and Adam Lerer.
\newblock Automatic differentiation in pytorch.
\newblock 2017.

\bibitem[Poole et~al.(2023)Poole, Jain, Barron, and Mildenhall]{poole2023dreamfusion}
Ben Poole, Ajay Jain, Jonathan~T. Barron, and Ben Mildenhall.
\newblock Dreamfusion: Text-to-3d using 2d diffusion.
\newblock In \emph{The Eleventh International Conference on Learning Representations}, 2023.

\bibitem[Qin et~al.(2024)Qin, Li, Zhou, Wang, and Pfister]{qin2024langsplat}
Minghan Qin, Wanhua Li, Jiawei Zhou, Haoqian Wang, and Hanspeter Pfister.
\newblock Langsplat: 3d language gaussian splatting.
\newblock In \emph{Proceedings of the IEEE/CVF Conference on Computer Vision and Pattern Recognition}, pages 20051--20060, 2024.

\bibitem[Radford et~al.(2021)Radford, Kim, Hallacy, Ramesh, Goh, Agarwal, Sastry, Askell, Mishkin, Clark, et~al.]{radford2021learning}
Alec Radford, Jong~Wook Kim, Chris Hallacy, Aditya Ramesh, Gabriel Goh, Sandhini Agarwal, Girish Sastry, Amanda Askell, Pamela Mishkin, Jack Clark, et~al.
\newblock Learning transferable visual models from natural language supervision.
\newblock In \emph{International conference on machine learning}, pages 8748--8763. PMLR, 2021.

\bibitem[Ranftl et~al.(2021)Ranftl, Bochkovskiy, and Koltun]{ranftl2021vision}
Ren{\'e} Ranftl, Alexey Bochkovskiy, and Vladlen Koltun.
\newblock Vision transformers for dense prediction.
\newblock In \emph{Proceedings of the IEEE/CVF international conference on computer vision}, pages 12179--12188, 2021.

\bibitem[Rombach et~al.(2022)Rombach, Blattmann, Lorenz, Esser, and Ommer]{rombach2022high}
Robin Rombach, Andreas Blattmann, Dominik Lorenz, Patrick Esser, and Bj{\"o}rn Ommer.
\newblock High-resolution image synthesis with latent diffusion models.
\newblock In \emph{Proceedings of the IEEE/CVF conference on computer vision and pattern recognition}, pages 10684--10695, 2022.

\bibitem[Schonberger and Frahm(2016)]{schonberger2016structure}
Johannes~L Schonberger and Jan-Michael Frahm.
\newblock Structure-from-motion revisited.
\newblock In \emph{Proceedings of the IEEE conference on computer vision and pattern recognition}, pages 4104--4113, 2016.

\bibitem[Song et~al.(2021)Song, Meng, and Ermon]{song2021denoising}
Jiaming Song, Chenlin Meng, and Stefano Ermon.
\newblock Denoising diffusion implicit models.
\newblock In \emph{International Conference on Learning Representations}, 2021.

\bibitem[Suvorov et~al.(2022)Suvorov, Logacheva, Mashikhin, Remizova, Ashukha, Silvestrov, Kong, Goka, Park, and Lempitsky]{suvorov2022resolution}
Roman Suvorov, Elizaveta Logacheva, Anton Mashikhin, Anastasia Remizova, Arsenii Ashukha, Aleksei Silvestrov, Naejin Kong, Harshith Goka, Kiwoong Park, and Victor Lempitsky.
\newblock Resolution-robust large mask inpainting with fourier convolutions.
\newblock In \emph{Proceedings of the IEEE/CVF winter conference on applications of computer vision}, pages 2149--2159, 2022.

\bibitem[Wang et~al.(2024{\natexlab{a}})Wang, Zhang, Abboud, and S{\"u}sstrunk]{wang2024innerf360}
Dongqing Wang, Tong Zhang, Alaa Abboud, and Sabine S{\"u}sstrunk.
\newblock Innerf360: Text-guided 3d-consistent object inpainting on 360-degree neural radiance fields.
\newblock In \emph{Proceedings of the IEEE/CVF Conference on Computer Vision and Pattern Recognition}, pages 12677--12686, 2024{\natexlab{a}}.

\bibitem[Wang et~al.(2021)Wang, Liu, Liu, Theobalt, Komura, and Wang]{wang2021neus}
Peng Wang, Lingjie Liu, Yuan Liu, Christian Theobalt, Taku Komura, and Wenping Wang.
\newblock Neus: Learning neural implicit surfaces by volume rendering for multi-view reconstruction.
\newblock \emph{arXiv preprint arXiv:2106.10689}, 2021.

\bibitem[Wang et~al.(2024{\natexlab{b}})Wang, Wu, Zhang, and Xu]{wang2024gscream}
Yuxin Wang, Qianyi Wu, Guofeng Zhang, and Dan Xu.
\newblock Learning 3d geometry and feature consistent gaussian splatting for object removal.
\newblock In \emph{European Conference on Computer Vision}, pages 1--17. Springer, 2024{\natexlab{b}}.

\bibitem[Weber et~al.(2024)Weber, Holynski, Jampani, Saxena, Snavely, Kar, and Kanazawa]{weber2024nerfiller}
Ethan Weber, Aleksander Holynski, Varun Jampani, Saurabh Saxena, Noah Snavely, Abhishek Kar, and Angjoo Kanazawa.
\newblock Nerfiller: Completing scenes via generative 3d inpainting.
\newblock In \emph{Proceedings of the IEEE/CVF Conference on Computer Vision and Pattern Recognition}, pages 20731--20741, 2024.

\bibitem[Wu et~al.(2024)Wu, Bian, Li, Wang, Reid, Torr, and Prisacariu]{wu2024gaussctrl}
Jing Wu, Jia-Wang Bian, Xinghui Li, Guangrun Wang, Ian Reid, Philip Torr, and Victor~Adrian Prisacariu.
\newblock Gaussctrl: Multi-view consistent text-driven 3d gaussian splatting editing.
\newblock In \emph{European Conference on Computer Vision}, pages 55--71. Springer, 2024.

\bibitem[Wu et~al.(2023)Wu, Ge, Wang, Lei, Gu, Shi, Hsu, Shan, Qie, and Shou]{wu2023tune}
Jay~Zhangjie Wu, Yixiao Ge, Xintao Wang, Stan~Weixian Lei, Yuchao Gu, Yufei Shi, Wynne Hsu, Ying Shan, Xiaohu Qie, and Mike~Zheng Shou.
\newblock Tune-a-video: One-shot tuning of image diffusion models for text-to-video generation.
\newblock In \emph{Proceedings of the IEEE/CVF International Conference on Computer Vision}, pages 7623--7633, 2023.

\bibitem[Yang et~al.(2021)Yang, Lamdouar, Lu, Zisserman, and Xie]{yang2021self}
Charig Yang, Hala Lamdouar, Erika Lu, Andrew Zisserman, and Weidi Xie.
\newblock Self-supervised video object segmentation by motion grouping.
\newblock In \emph{Proceedings of the IEEE/CVF International Conference on Computer Vision}, pages 7177--7188, 2021.

\bibitem[Zhang et~al.(2023)Zhang, Rao, and Agrawala]{zhang2023adding}
Lvmin Zhang, Anyi Rao, and Maneesh Agrawala.
\newblock Adding conditional control to text-to-image diffusion models.
\newblock In \emph{Proceedings of the IEEE/CVF International Conference on Computer Vision}, pages 3836--3847, 2023.

\bibitem[Zhang et~al.(2018)Zhang, Isola, Efros, Shechtman, and Wang]{zhang2018unreasonable}
Richard Zhang, Phillip Isola, Alexei~A Efros, Eli Shechtman, and Oliver Wang.
\newblock The unreasonable effectiveness of deep features as a perceptual metric.
\newblock In \emph{Proceedings of the IEEE conference on computer vision and pattern recognition}, pages 586--595, 2018.

\bibitem[Zhou et~al.(2023)Zhou, Li, Chan, and Loy]{zhou2023propainter}
Shangchen Zhou, Chongyi Li, Kelvin~CK Chan, and Chen~Change Loy.
\newblock Propainter: Improving propagation and transformer for video inpainting.
\newblock In \emph{Proceedings of the IEEE/CVF International Conference on Computer Vision}, pages 10477--10486, 2023.

\bibitem[Zhou et~al.(2024)Zhou, Chang, Jiang, Fan, Zhu, Xu, Chari, You, Wang, and Kadambi]{zhou2024feature}
Shijie Zhou, Haoran Chang, Sicheng Jiang, Zhiwen Fan, Zehao Zhu, Dejia Xu, Pradyumna Chari, Suya You, Zhangyang Wang, and Achuta Kadambi.
\newblock Feature 3dgs: Supercharging 3d gaussian splatting to enable distilled feature fields.
\newblock In \emph{Proceedings of the IEEE/CVF Conference on Computer Vision and Pattern Recognition}, pages 21676--21685, 2024.

\bibitem[Zhou et~al.(2021)Zhou, Barnes, Shechtman, and Amirghodsi]{zhou2021transfill}
Yuqian Zhou, Connelly Barnes, Eli Shechtman, and Sohrab Amirghodsi.
\newblock Transfill: Reference-guided image inpainting by merging multiple color and spatial transformations.
\newblock In \emph{Proceedings of the IEEE/CVF conference on computer vision and pattern recognition}, pages 2266--2276, 2021.

\bibitem[Zhuang et~al.(2023)Zhuang, Zeng, Liu, Yuan, and Chen]{zhuang2023task}
Junhao Zhuang, Yanhong Zeng, Wenran Liu, Chun Yuan, and Kai Chen.
\newblock A task is worth one word: Learning with task prompts for high-quality versatile image inpainting.
\newblock \emph{arXiv preprint arXiv:2312.03594}, 2023.

\bibitem[Zwicker et~al.(2001)Zwicker, Pfister, Van~Baar, and Gross]{zwicker2001ewa}
Matthias Zwicker, Hanspeter Pfister, Jeroen Van~Baar, and Markus Gross.
\newblock Ewa volume splatting.
\newblock In \emph{Proceedings Visualization, 2001. VIS'01.}, pages 29--538. IEEE, 2001.

\end{thebibliography}
